\documentclass[runningheads]{llncs}

\usepackage{hyperref}

\begin{document}

\title{Cem Mil Podcasts: A Spoken Portuguese Document Corpus For Multi-modal, Multi-lingual and Multi-Dialect Information Access Research}
\titlerunning{Cem Mil Podcasts}  

\author{
Ekaterina Garmash\inst{1} 
\and
Edgar Tanaka\inst{1} 
\and
Ann Clifton\inst{1}  
\and
Joana Correia\inst{1}
\and
Sharmistha Jat  
\and
Winstead Zhu\inst{1}  
\and
Rosie Jones\inst{1} 
\and
Jussi Karlgren\inst{2} 
}

\institute{Spotify
\and
Silo AI
}

\authorrunning{E. Garmash et al.}


\maketitle

\begin{abstract}
In this paper we describe the Portuguese-language podcast dataset we  have released for academic research purposes. We give an overview of how the data was sampled,  descriptive statistics over the collection, as well as  information about the distribution over Brazilian and Portuguese dialects.

We give results from experiments on multi-lingual summarization, showing that summarizing podcast transcripts can be performed well by a system supporting both English and Portuguese. We also show experiments on Portuguese podcast genre classification using text metadata. 
Combining this collection with previously released English-language collection opens up the potential for multi-modal, multi-lingual and multi-dialect podcast information access research.
\keywords{Dataset  \and Podcast  \and Spoken audio \and Speech retrieval \and Multi-modal \and Summarization}  
\end{abstract}  
\section{Introduction}
Podcasts, a new and emergent spoken mass communication medium, come in many formats and levels of formality. Podcasts are typically produced as topically or stylistically consistent \textit{shows} that consist of \textit{episodes}, which are published serially with some regularity over time. 
Podcast consumption has been growing rapidly \cite{whitner2020meteoric} and podcasts have in the past years become a topic of interest for research in speech and language technology, linguistics, information access technology, and media studies. 

Podcast shows can be educational, journalistic, or fictional; formal or informal; conversational or monologic; and vary over type, style, form, and topic. Podcast material appears to differ in several aspects from other types of recorded speech or textual material \cite{karlgren2022lexical}. This breadth of variation and contrast with other collections of human language motivates using podcasts for research both to develop infrastructure and tools for podcast distribution and consumption as well as to broaden the scope of the general study of human communicative behavior. 

Academic research on podcasts requires availability of open-source and representative datasets. 
The currently largest available collection is Spotify's English language podcast dataset \cite{clifton2020hundredthousand} which differs from other collections of English language data in that it is orders of magnitude larger than previous collections of spoken language and contains a rich variety of genres, subject matter, speaking styles, and structural formats.

Podcasts are available in a wide variety of languages. For example, the Anchor podcast creation app is available in thirty-five languages \cite{anchorUILanguages} with many more podcasters creating podcasts in other languages. 
However, the Spotify English language podcast dataset is, as are most available datasets, composed entirely of English-language material. Research on English alone risks results to be biased towards linguistic specifics of one language and cultural arenas where English mostly is produced and used. 

To address this source of bias in podcast research, we have compiled a complementary dataset of Portuguese-language material, with comparable size and breadth using the same general methodology as in \cite{clifton2020hundredthousand}. This dataset contains metadata similar to what is provided together with the Spotify English-language dataset, which can be used as proxy labels for supervised learning tasks, such as creator-provided textual episode descriptions for the summarization task and names of show publishers for authorship attribution tasks.  

The Portuguese language is the sixth most-spoken language in the world, with 250 million native speakers, in many cultural areas, and with 24 million more L2 speakers \cite{portLangWikipedia}, and the Lusophone markets are, taken together in the top ten of the world by GDP \cite{GDP}. This motivates a growing interest in working on language technology for Portuguese: as shown not least by the recent release of a large transformer model based on written Portuguese~\cite{rodrigues2023advancing}. 

In order to facilitate research on spoken Portuguese in general, and podcasts in Portuguese more specifically, we now make available a podcast dataset consisting of 123,054 podcast episodes in Portuguese from 16,131 shows, encompassing more than 76,000 hours of speech audio.

We know of no previous large-scale study of Portuguese language podcasts.  
There have been smaller-scale studies of Portuguese podcasts \cite{antunes2020examining}. Morais et al \cite{morais2021audience} surveyed 566 Brazilian podcast listeners, and found that they listen to podcasts across a range of topic areas, and that they like podcasts to convey information that is complementary to the information found in other media formats, similarly to what has been found in other linguistic and cultural areas. 




\section{Dataset construction} \label{sec:dataset_construct}

To construct the Portuguese dataset we followed a procedure 
patterned on the approach 
used 
Clifton et al.  \cite{clifton2020hundredthousand}.  
to build the English-language podcast dataset.

From a fairly comprehensive list of Portuguese podcasts 
we selected data based on the following filters:
\begin{itemize}
\item The language of the show as given by the podcast creator in the show metadata specification must be Portuguese (pt-BR or pt-PT). 
\item The language of the episode description must be identified as Portuguese using the langid Python package \cite{lui-baldwin-2012-langid}.
\item We only selected episodes published between September 9, 2019 and March 31, 2022.
\item The episode must have more than 50\% of speech over its duration. A proprietary speech detection algorithm was used here to filter out podcasts which contain mostly music, white noise or ambient sounds, rather than speech. 
\end{itemize}

From the list of filtered candidates, episodes were randomly selected to obtain just over 150~000 individual items.

The next step was to transcribe this set of episodes using Azure's speech-to-text service\footnote{https://docs.microsoft.com/en-us/azure/cognitive-services/speech-service/index-speech-to-text}. One of the parameters of this service was the target language-variant which we set to either pt-PT (Portuguese from Portugal) or pt-BR (Portuguese from Brazil) according to the following metadata in this order of precedence: creator-provided language code given in the \textit{show} metadata is either `pt-PT', the creator-provided language code given in the \textit{episode} metadata, the \textit{show's} country of origin is `pt' or `br'. If no metadata is set to either pt-PT or pt-BR, we fall back to `pt-BR' because the number of podcast creators in Brazil is larger.

Despite sending the entire pool of approximately 150~000 episodes to transcription, some of them failed to be transcribed. In the end, 114,387 episodes were transcribed using `pt-BR' and 8,667 were transcribed using `pt-PT' as target language. Examples of each are shown in Figure~\ref{fig:transcript}. Manual inspection reveals that some classification errors between `pt-PT' and `pt-BR' remain: words such as ``\textit{legal}'' (Brazilian for ``cool'') or the Brazilian-only pronoun ``cê'' appear in the pt-PT set.

\begin{figure}
\centering
\begin{tabular}{|lp{0.9\textwidth}|}
\hline
pt-PT & {\small
\textit{Olá \textbf{és} curioso sobre o que se passa no mundo, gosta de saber o que afetou diversosecossistemas e como podes ajudá-los. E esta ao \textbf{sítio} certo, não podcast e como escola São Pedro do Sul, \textbf{podes} encontrar informação sobre o ambiente, novidades relacionadas com o nosso planeta e muitas curiosidades, incluindo as diversas medidas que podes aplicar para ajudares a proteger a natureza. \textbf{Serás} capaz de aprofundar os teus conhecimentos sobre diversos temas, desde alterações climáticas até ao ruído. A \textbf{equipa} de 2 ou 3 vezes por mês que vai contar tudo é composta pelo Miguel Almeida, Rodrigo Cardoso. Tiago Rocha e Filipe Correia. Para não perder nada, subscreve. Já o nosso podcast na tua plataforma preferida e não te esqueças após cada episódio, Podes sempre visitar o nosso website para saberes mais.} } \\
 \hline
pt-BR &
{\small
\textit{ Olá \textbf{você} que é nosso ouvinte do podcast de arte saúde o fiba hoje estaremos nossa segunda entrevista e contaremos com a presença de uma convidada mais do que especial? Ela estéfane psicóloga e arte terapeuta. Oi eu sou estefani eu sou psicóloga formada pela universidade de Passo Fundo com especialização em arteterapia também pela universidade de Passo Fundo e recentemente eu encontrei uma ponte entre a psicologia e a arte terapia através de uma especialização em psicologia clínica e um. Indo pela fam acne Porto Alegre eu \textbf{tô} muito feliz com o convite da área de saúde para falar um pouquinho sobre esse assunto que deixa o meu coração tão quentinho que é arte terapia é sempre 
…}}
 \\
 \hline
\end{tabular}
    \caption{Example transcripts for the two target language varieties of Portuguese. The pt-PT example was extracted from the show ``EESPS: Podcast sobre o ambiente''. The pt-BR was extracted from the show ``Arte e Saúde - UFBA''.}
    \label{fig:transcript}
\end{figure}

\subsection{Dataset schema}

For each episode, we provide the audio file, the transcription of this audio file and the associated metadata. The following metadata is provided:
\begin{itemize}
\item \textit{show\_uri}: URI for the show 
\item \textit{show\_name}: Name of the show (e.g. ``Hoje no TecMundo Podcast'').
\item \textit{show\_description}: Description of the show provided by podcast creator (e.g. ``O Hoje no TecMundo é o tradicional programa diário do TecMundo no YouTube...”)
\item \textit{publisher}: Publisher of the show (e.g. Hoje no TecMundo - Podcast).
\item \textit{language}: Language of the show in in BCP 47 format (e.g. pt-BR).
\item \textit{rss\_link}: URL of the show's RSS feed (e.g. https://anchor.fm/s/11c4550c/podcast/rss). 
\item \textit{episode\_uri}: URI for the episode 
\item \textit{episode\_name}: Name of the episode. (e.g. "Hoje no TecMundo 17/01/2020 – Preço do Galaxy Fold no Brasil, imagens do Huawei P40 Pro")
\item \textit{episode\_description}: Description of the episode (e.g. “No programa de hoje, falamos do preço caríssimo do Galaxy Fold no Brasil, a Google ...”).
\item \textit{duration}: duration of the episode in minutes (e.g. 9.113833333333334).
\item \textit{show\_filename\_prefix}: Filename path for the show 
\item \textit{episode\_filename\_prefix}: Filename of the episode file 
\item \textit{show\_category}: The genre of the show extracted from the \textit{itunes:category} tag in the show's RSS feed.
\end{itemize}

\subsection{Access to the dataset}

The Portuguese language podcast dataset is available for non-commercial research purposes in the same way and under the same agreement as the English language podcast dataset. The English language podcast dataset has been most notably used in shared tasks on segment retrieval and summarization in TREC 2020 \cite{jones2020trec} and TREC 2021 \cite{karlgren2021trec} but is also currently used for many other research purposes such as document segmentation or dialogue modelling and there are several annotations and enrichments such as a search index, human assessments, and precomputed audio features available for the English language section. We expect that this extension will broaden the scope of research and lower the threshold to apply methods to more than one language. We welcome contributions to further enrich the dataset through annotations of various kinds. 
 
 To request the dataset, please go to \url{https://podcastsdataset.byspotify.com/} and follow the instructions.


\section{Descriptive statistics and comparison to English podcast dataset}
\label{sec:data_stats}

In Table \ref{tab:pt-general-stats}, we give some descriptive statistics for the Portuguese dataset and compare them to 
the English-language dataset released by 
\cite{clifton2020hundredthousand}
to demonstrate that they are of comparable size (in terms of number of episodes) and quality.
We note however that the English set turns out to have a slightly larger diversity of shows: the show-to-episode ratio for English is 17\%, while it is 13\% for Portuguese. This is an expected consequence of there being more podcast shows in English than in Portuguese, and thus selecting approximately the same number of episodes will yield slightly more episodes per show for Portuguese than for English. The distribution of episode durations also differ: the Portuguese episodes tend to be longer on average and the distribution is skewed to the right. As a consequence, the average number of words per episodes is also higher for Portuguese. This observation entails, in particular, that development of machine learning models of podcast understanding may be more challenging for Portuguese, since its input size will be larger. To illustrate this we provide a case study on episode summarization in (Section~\ref{sec:summarization}).


Given that the general dataset construction procedure
is similar
for both datasets, and that the analyzed samples are of comparable sizes, the detected differences in the distributions of various features support our original claim that linguistically diverse data is necessary to avoid biased conclusions in podcast research.


\begin{table}[htbp]
\centering
\caption{Descriptive statistics for the Portuguese Language Podcast Dataset, with corresponding data for the Spotify English Language Podcast Dataset given as comparison.}
\begin{tabular}{|r|r|r|}
\hline
     & English & Portuguese  \\
\hline
Number of episodes     & 105 360 & 123 054\\
    Number of shows     &      18 376  &  16 131 \\
\hline
    Average episode duration (minutes) & 33.8 & 37.3 \\
- 25\% & 13.6 & 10.9  \\ 
- 50\% & 31.6 & 31.2   \\ 
- 75\% & 50.4 & 55.0   \\ 
- max   & 305 & 695  \\ 
\hline
    Average number of words per transcript  &  5 726 &  9 539  \\
- 25\% & 2 036 & 2 203                  \\ 
- 50\% & 5 204 & 6 746                  \\ 
- 75\% & 8 672 & 13 693                 \\ 
- max  & 43 504 & 205 163  \\ 
\hline

\hline
\end{tabular}
\label{tab:pt-general-stats}
\end{table}

\section{Podcast genre prediction case study}
\label{sec:genre_classif}
Besides being multi-modal, our dataset comes with rich metadata annotation. We demonstrate the usefulness of metadata by benchmarking the task of podcast genre prediction. Podcast genres are essential when understanding user taste in order to recommend related listening experiences. Intrinsically, the genre is a characterization of the podcast's content as a whole and could therefore be inferred based on the raw podcast data (audio or text). However, we show in our experiments that the metadata we provide in the dataset could be sufficient for the task, which is beneficial from a practical perspective. Specifically, we run genre prediction experiments where input is restricted to episode names and a short episode description (summary),  which is often provided by creators and which does not need to be generated separately.

\subsection{Genre Prediction Experiment Setup}
As target labels, we use creator-provided genre labels located in the \textit{show\_category} column of the metadata provided. The taxonomy consists of 19 genre labels. In Table \ref{tab:pt-genre}, we can see the distribution of episodes per genre. The top 5 genres account for 69\% of all episodes. We also note that \textit{Business}, \textit{Education}, \textit{Sports}, and \textit{Comedy} are all within the top 5 genres. 
This is similar to what was reported for Spotify's English Language dataset \cite{clifton2020hundredthousand}. Please note that the number of episodes in the genre prediction task is a subset of the total number of episodes in the dataset. We provide the train and test split of the dataset used in the prediction task.

We frame genre prediction as a classification task. Predictions are made per episode. As input, we use episode names and episode descriptions, individually and combined (see Results for an ablation study). We use a multi-class support vector machine for the classification task using \textit{sklearn}'s  \cite{sklearn_api} \textit{SVC} class implementation with default hyperparameter settings. The episode-name and episode-description text inputs are pre-processed using \textit{bert-base-multi-lingual-uncased} \cite{devlin-etal-2019-bert} to generate 768-sized embeddings.  Train and test splits are created using an 80:20 split, making sure that the split is partitioned by show URIs in order to avoid any information leakage between test and train splits (since episodes from the same show have a common genre). 

We consider three experimental conditions: (1) using episode name input only; (2) using episode description input only; (3) using both episode name and description as input. We report precision, recall, and F1 of test prediction results by genre, as well as aggregate accuracy and macro \& weighted averages of these metrics over all the genres.

\begin{table*}[th]
\centering
\caption{Portuguese Podcasts dataset: number of episodes per genre}

\begin{tabular}{c|c|c}
\hline
\textbf{S.No.}          
& \textbf{Genre}           
& \textbf{Number of Episodes} \\ 
   &                      & \textbf{in Portuguese Dataset} \\ \hline
1 & Business                 
& 26~915                   \\ \hline
2 & Education                
& 23~541                   \\ \hline

3 & Sports                   
                               &13~422                   \\ \hline
4 & Comedy                   
                               &11~089                   \\ \hline

5 & Arts                     
                               &9~799                    \\ \hline
6 & TV \& Film               
                               &5~445                    \\ \hline

7 & Science                  
                              & 5~371                    \\ \hline
8 & Music                    
                               & 3~916                    \\ \hline

9 & Technology               
                               & 3~186                    \\ \hline
10 & Society \& Culture       
                               & 3~172                    \\ \hline

11 & Kids \& Family           
                              & 3~026                    \\ \hline
12 & Leisure                  
                              & 2~647                    \\ \hline

13 & Health \& Fitness        
                              & 2~337                    \\ \hline
14 & History                  
                              & 2~213                    \\ \hline

15 & Fiction                  
                              & 2~017                    \\ \hline
16 & True Crime               
                              & 1~748                    \\ \hline

17 & News                     
                              & 1~097                    \\ \hline
18 & Religion \& Spirituality 
                              & 1~033                    \\ \hline

19 & Government               
                              & 531                     \\ \hline\hline
\multicolumn{2}{c|}{Total}                     
                              & 122~5273                     \\ \hline
\end{tabular}
\label{tab:pt-genre}
\end{table*}

\subsection{Genre Prediction Experiment Results}
All the results are summarized in Table~\ref{tab:genre_result}. Firstly, we see that for the top genres (that we identified in  \autoref{tab:pt-genre}), the F1 scores are substantially above 0.5, which indicates that given enough data, just the name and the description are sufficient to identify a podcast genre correctly. Second, as intuitively expected, the description of episodes contributes more to prediction quality. However, combining them both yields the best result. As for the rest of the genres, the prediction quality is much lower, which could be explained by the low data support; \textit{News} e.g. has only 1097 episodes compared to 26,915 in the \textit{Business} genre.


\begin{table*}[th]
\centering
\caption{Results of the genre prediction classification experiments: precision, recall, and F1 are reported by genre and as averages across genres. The top most popular genres (according to Table~\ref{tab:pt-genre}) are marked in boldface. We observe that the popular genres have an F1 score above 0.5, showing that metadata alone can detect the genre of a given episode. Although episode description features dominate the Genre prediction accuracy, adding the episode name features helps improve the genre prediction metrics in most genres.}
\begin{tabular}{l|c|c|c|c|c|c|c|c|c}
\hline
\textbf{Genre} & \multicolumn{3}{c|}{Name} & \multicolumn{3}{c|}{Description} & \multicolumn{3}{c}{Name \& Description} \\
\hline
& Prec. & Rec. & F1 & Prec. & Rec. & F1 & Prec. & Rec. & F1 \\
\hline
Arts   & 0.31  &   0.3  &   0.31   &  0.34  &   0.38   &  0.36  &  0.42  &   0.47   &  0.44   \\  
\textbf{Business} & 0.44 & 0.68 & 0.54 &   0.72  & 0.63 & 0.57   &  0.75    & 0.65\\ 
\textbf{Comedy}   &  0.38 & 0.32 & 0.35  & 0.54     & 0.59  & 0.56 & 0.53   &  0.57   &  0.55  \\   
\textbf{Education}   &  0.48 & 0.66 & 0.56 & 0.5 & 0.72 & 0.59  & 0.52 &   0.73  &   0.61 \\  
Fiction    &   0.08 & 0.03 & 0.04 & 0.27 & 0.03 & 0.05 & 0.16 &    0.07  & 0.1   \\  
Government   &   0.0   &  0.0  &   0.0  &   0.0   &  0.0  &   0.0 &   0.0   &  0.0  &   0.0   \\ 
Health and Fitness   &   0.0   &  0.0  &   0.0  &   0.0   &  0.0  &   0.0 &   0.0   &  0.0  &  0.0     \\
History  &     0.22 & 0.05 & 0.08 & 0.36 & 0.16 & 0.22 & 0.39   &  0.11  &   0.17    \\
Kids and Family   &   0.33 & 0.08 & 0.13 & 0.47 & 0.07 & 0.12  & 0.51   &  0.09 &     0.15    \\
Leisure   &    0.07 & 0.06 & 0.07 & 0.23 & 0.14 & 0.18 & 0.21 &     0.13 &     0.16    \\
Music   &   0.45 & 0.3 & 0.36 & 0.64 & 0.48 & 0.55 &  0.65 &     0.55  &   0.6   \\
News   &   0.0  &   0.0  &   0.0 &  0.0  &   0.0  &   0.0 &  0.0  &   0.0  &   0.0       \\
Religion and Spirituality    &  0.0 & 0.0 & 0.0 & 0.67 & 0.03 & 0.05 & 0.75   &  0.02  &   0.04     \\
Science  &    0.39 & 0.08 & 0.13 & 0.41 & 0.11 & 0.17 &  0.47  &   0.13    &  0.21     \\
Society and Culture   &   0.0 & 0.0 & 0.0 & 0.76 & 0.04 & 0.08 &    
 0.52  &   0.02  &   0.04 \\
\textbf{Sports}   &   0.52 & 0.67 & 0.58 & 0.81 & 0.79 & 0.8 & 0.76   &  0.83 &   0.79    \\
TV and Film  &    0.38 & 0.27 & 0.32 & 0.49 & 0.42 & 0.46    & 0.5 &     0.4  &     0.45    \\
Technology  &    0.14 & 0.02 & 0.03 & 0.46 & 0.09 & 0.15 & 0.52 &     0.09  &     0.16  \\
True Crime   &   0.37 & 0.10 & 0.16 & 0.51 & 0.34 & 0.41 & 0.48 &     0.32  &   0.38     \\
\hline
macro average & 0.23 &    0.18  &   0.18  & 0.4   &  0.25  &   0.27 & 0.4 & 0.26 & 0.27 \\
weighted average & 0.39  & 0.44 & 0.39 & 0.52  & 0.53  &   0.49 & 0.53  &   0.55   & 0.51 \\\hline
Accuracy & \multicolumn{3}{c|}{0.44} & \multicolumn{3}{c|}{0.53} & \multicolumn{3}{c}{0.55}\\
\hline
\end{tabular}
\label{tab:genre_result}
\end{table*}

\section{Episode Summarization Case Study}
\label{sec:summarization}
In this section we present a case study of one of the possible machine learning applications of our dataset: episode summarization. 
Automatic Text Summarization is the task of taking a source document as input and producing a much shorter version of it while preserving the most important pieces of information \cite{comprehensive-survey}. One of the differences between podcast episode summarization  and the many other domains where summarization is applied is the extensive length of the podcast input document, which poses a challenge since currently most neural network-based approaches are limited in terms of input size. Moreover, as we saw in Table~\ref{tab:pt-general-stats}, Portuguese episodes tend to be longer than the English ones, which suggests a research question: is episode summarization more challenging for Portuguese? We address this question by conducting a series of machine learning experiments to train a transcript-based summarization model, and compare the resulting quality for English (based on \cite{clifton2020hundredthousand}) and Portuguese (based on the dataset from this paper).

\subsection{Data Preparation for Summarization}
As input to the summarization model, we use the automatically generated episode transcripts described in Section \ref{sec:dataset_construct}. We treat creator-provided episode descriptions as the summaries and train the model to generate them. 

We further clean the data using the following filters:
\begin{itemize}
\item We remove episodes with repeated descriptions (any description used in more than one episode). We applied a TF-IDF vectorization of the descriptions which were compared to each other using the cosine distance. Any data points with too similar descriptions (threshold 95\%) were filtered out.
\item We remove episodes where the episode description is too similar to the show description (threshold 95\%).
\item We remove any email addresses or URLs from episode descriptions as we did not want our trained models to hallucinate such information in the generated summaries.
\item We remove episodes where the creator descriptions are either too long or too short with the boundary conditions set to between 10 and 1300 characters. 
\item We remove \textit{boilerplate} content from episode descriptions. Briefly speaking, boilerplate is any extraneous content which does not describe the episode in natural language text. Common cases of boilerplate in podcasts are advertisements and promotional content for social media \cite{reddy2021detecting}. We train a sentence-level binary classifier for that, based on a small manually annotated dataset and fine-tune a pretrained language model (\textit{bert-base-cased} for English and \textit{bert-base-multi-lingual-cased} for Portuguese).
\end{itemize}

After applying the filters above, we split the remaining data into 3 parts: train (90\%), dev (5\%) and test (5\%). The split was a per-show partitioning of the data. Tables~\ref{tab:summ_split} contains statistics of the resulting splits.

\begin{table}[h]
\centering
\caption{Experimental data for summarization: split size by number of episodes.}
{\small 
\begin{tabular}{llrr}
\hline
 & \textbf{ratio} & \textbf{EN} & \textbf{PT} \\ 
 \hline
train & 90\% & 80~895 & 90~859 \\ 
dev & 5\% & 4~503 & 5~073 \\ 
test & 5\% & 4~511 & 5~058 \\ \hline
\end{tabular}
}
\label{tab:summ_split}
\end{table}

\subsection{Models for Summarizatiom}
For baselines, we follow \cite{clifton2020hundredthousand} \cite{karlbom2020abstractive} \cite{jones2020trec} \cite{karlgren2021trec} and use the \textbf{first minute} transcript and \textbf{TextRank} \cite{mihalcea2004textrank}, 
a graph-based model which can be used as an unsupervised method to extract both keywords or key sentences. Both baselines are extractive summarization methods and do not require any training.

We run machine learning experiments with \textbf{MBART} \cite{mbart-50}, a multi-lingual version of BART \cite{lewis2019bart}. We chose to use the MBART-50\cite{liu2020multilingual} model because it has been pre-trained in 50 languages (including Portuguese and English) and also because it is an encoder-decoder model, i.e. capable of generating text. Additionally, we replace MBART's original full attention mechanism with the one of Longformer \cite{beltagy2020longformer} -- we refer to this model variation as \textbf{LongMBART}.  LongMBART's linear attention allows to increase the input size limit from 512 tokens to 4096 tokens. Our hypothesis is that passing more information (i.e. more transcript text) to the model would lead to higher scores. 

For both MBART and LongMBART, we consider three experimental conditions: (1) unfinetuned base model; (2) finetuned on the language of the test set (``monolingual fine-tuning''); (3) finetuned on both the English and Portuguese training set (``bilingual fine-tuning''). The fine-tuning was set to early stop once the ROUGE-2 \cite{lin-2004-rouge}  score didn't improve after 3 validation checkpoints. 


\subsection{Results for Summarization}
We report ROUGE-1 (unigram overlap), ROUGE-2 (bigram overlap), ROUGE-L (longest matching sequence of words)\cite{lin-2004-rouge} scores 
on the test set in Table~\ref{tab:summ_exp}. First, we see that both languages obtain the highest scores for the same types of model (fine-tuned MBART). Second, we see no 
 substantial difference between the best scores of English and Portuguese, contrary to our hypothesis that Portuguese should be more challenging given its larger input size. Finally, we see that using LongMBART as base model does not result in better summarization.

To sum up, having two comparable podcast datasets for different languages with intrinsically different data distribution allowed us to test a series of hypotheses: effect of linguistic and cultural specifics on the quality of machine learning-based summarization model, effect of size of input, effect of multi-lingual fine-tuning.

\begin{table}[h]
\centering
\caption{ROUGE-1, ROUGE-2 and ROUGE-L F1 scores for test set of 4511 English-language podcast episodes and test set of 5073 Portuguese-language podcast episodes. In bold, the top two highest ROUGE scores. } 
\begin{tabular}{lrrr|rrr}
\hline
& \multicolumn{3}{c|}{\textbf{EN}} & \multicolumn{3}{c}{\textbf{PT}} \\
\textbf{}                  & \multicolumn{1}{l}{\textbf{R1}} & \multicolumn{1}{l}{\textbf{R2}} & \multicolumn{1}{l|}{\textbf{RL}} & \multicolumn{1}{l}{\textbf{R1}} & \multicolumn{1}{l}{\textbf{R2}} & \multicolumn{1}{l}{\textbf{RL}}  \\ \hline
First Minute baseline                & 0.17                           & 0.03                            & 0.15   & 0.17                            & 0.03                            & 0.14                                         \\
TextRank Top 5 sentences             & 0.14                            & 0.02                            & 0.12   & 0.17                            & 0.01                            & 0.1                           \\
\hline
MBART unfinetuned                        & 0.16                            & 0.03                            & 0.14                & 0.16                            & 0.03                            & 0.13               \\
MBART finetuned monoling.        & \textbf{0.19}                            & \textbf{0.06}                           & \textbf{0.17}       & \textbf{0.19}    & \textbf{0.05}    & \textbf{0.16} \\
MBART finetuned biling.     & \textbf{0.19}                            & 0.05                            & \textbf{0.17}        & \textbf{0.18}                            & \textbf{0.05}                           & \textbf{0.16}            \\
\hline
LongMBART unfinetuned                    & 0.16                            & 0.03                            & 0.14                   & 0.11                            & 0.01                            & 0.1                  \\
LongMBART finetuned monoling.    & 0.03                            & 0.0                            & 0.03                       & 0.18                            &\textbf{0.05}                            & 0.16\\

LongMBART finetuned biling. & 0.18   & 0.05                            & 0.16  & 0.18                            & 0.05                            & 0.15   \\
\hline

\end{tabular}
\label{tab:summ_exp}
\end{table}

\section{Conclusions}
In this paper we presented a new dataset of Portuguese language podcasts, containing audio and transcript data, as well as rich metadata annotation, 
following the methodology used to put together the English language podcast dataset \cite{clifton2020hundredthousand}. 

Having a dataset in a language other than English, developed under the same methodology, allows for a more comprehensive and unbiased research of the podcast domain. From the point of view of machine learning research, linguistically diverse podcast data allows to study the effect of various input characteristics on the quality of the final model. In particular, in a case study presented in this paper, we show how our Portuguese dataset is used to train a podcast summarization model, and to compare it to an English summarization model, and thus test the effect of input language and size of input on the quality of the final model. Moreover, having data in multiple languages opens up possibility of research into multi-lingual machine learning models, widely adopted in other domains of natural language processing \cite{mbart-50,DBLP:journals/corr/abs-2010-11125,xue2021mt5}. Finally, metadata features which we supply in the dataset allow to train lightweight prediction models that do not need to take the full raw input. In another case study presented in this paper, we demonstrate how metadata can be used for genre prediction where we only used episode name and short description text as input.

Beyond the case studies presented in this paper, the new dataset can be used for many more experiments into multi-lingual and multi-modal (text and audio) machine learning models of podcasts. From a methodological point of view, we have demonstrated the reproducibility of the dataset construction procedure and plan to extend it to further languages.

\bibliographystyle{splncs04}
\bibliography{ptpodcastcorpus}
\end{document}